%% file: main.tex
\documentclass[10pt,twocolumn]{article}
\usepackage[margin=0.76in,top=0.82in,bottom=0.82in]{geometry}
\usepackage[T1]{fontenc}
\usepackage[utf8]{inputenc}
\usepackage{lmodern}
\usepackage{microtype}
\usepackage[round,authoryear]{natbib}
\usepackage{amsmath,amssymb,amsfonts}
\usepackage{booktabs,tabularx,array}
\usepackage{graphicx}
\usepackage{tikz}
\usetikzlibrary{arrows.meta,positioning,calc}
\usepackage{algorithm}
\usepackage{algpseudocode}
\usepackage{enumitem}
\usepackage{xcolor}
\usepackage{hyperref}
\usepackage{url}
\usepackage{caption}
\usepackage{listings}
\definecolor{deepblue}{RGB}{31,78,133}
\hypersetup{colorlinks=true,linkcolor=deepblue,citecolor=deepblue,urlcolor=deepblue,
 pdftitle={Adaptive Semantic Branching: Disagreement-Localized Cross-Trajectory Regeneration for Reasoning Candidate Expansion},
 pdfauthor={Jeong-Yoon Kim and Tae-Yoon Lee}}
\setlist[itemize]{leftmargin=*,itemsep=1pt,parsep=0pt,topsep=2pt}
\newcommand{\asb}{\textsc{ASB}}
\newcommand{\cove}{\operatorname{Cov}}

\title{\vspace{-0.4in}\bfseries Adaptive Semantic Branching:\\ Disagreement-Localized Cross-Trajectory Regeneration\\ for Reasoning Candidate Expansion}
\author{Jeong-Yoon Kim\\\small Department of Electronic Engineering\\\small Hanbat National University \and Tae-Yoon Lee\\\small Department of Electronic Engineering\\\small Hanbat National University}
\date{}
\begin{document}
\maketitle
\vspace{-0.14in}
\begin{abstract}
Reasoning trajectories generated for the same problem often arrive at conflicting answers. These disagreements reveal alternative intermediate decisions that can be reused to explore new solutions. We introduce \emph{Adaptive Semantic Branching} (\asb), a training-free inference procedure that turns cross-trajectory disagreement into targeted candidate generation. \asb\ first groups independently sampled trajectories by extracted final answer, selects a representative from each group, and aligns their reasoning steps with a monotonic dynamic program. A persistent-divergence objective identifies the point at which initially similar reasoning paths develop different continuations. For each selected directed pair, \asb\ retains the source trajectory's textual prefix, extracts a short reasoning window from a competing donor trajectory, and generates a fresh solution conditioned on both. A diversity-aware policy allocates up to three additional generations according to the answer-cluster structure. In a 500-problem MATH evaluation with Qwen2.5-7B-Instruct, \asb\ increases oracle candidate coverage from 445/500 (89.0\%) with seven initial samples to 455/500 (91.0\%), recovering correct candidates for 10 of the 55 initially uncovered problems. The method uses 576 additional generation calls across the evaluation set, averaging 8.152 calls per problem. We provide a precise specification of the branching operator, alignment objective, generation policy, and experimental configuration.
\end{abstract}

\section{Introduction}
Sampling multiple chains of thought is a standard approach to improve the breadth of language-model reasoning \cite{wang2023selfconsistency}. Different samples can follow distinct solution strategies, encounter different intermediate obstacles, and occasionally reach different final answers. Tree- and graph-structured reasoning methods make these alternatives explicit by expanding and recombining intermediate states \cite{yao2023tot,besta2024got}. An additional opportunity arises when a model has already produced several complete solutions: the relationships among these trajectories provide a structured signal for allocating subsequent generation.

Consider two completed solutions to the same mathematics problem. Their opening steps may encode a shared interpretation or calculation, while later steps pursue different substitutions, assumptions, or subgoals. A new candidate can draw on the stable initial reasoning of one solution while exploring an alternative local direction extracted from the other. This requires three concrete decisions: locating the transition between common and competing reasoning, selecting the information passed between trajectories, and choosing which directed exchanges receive a limited generation budget.

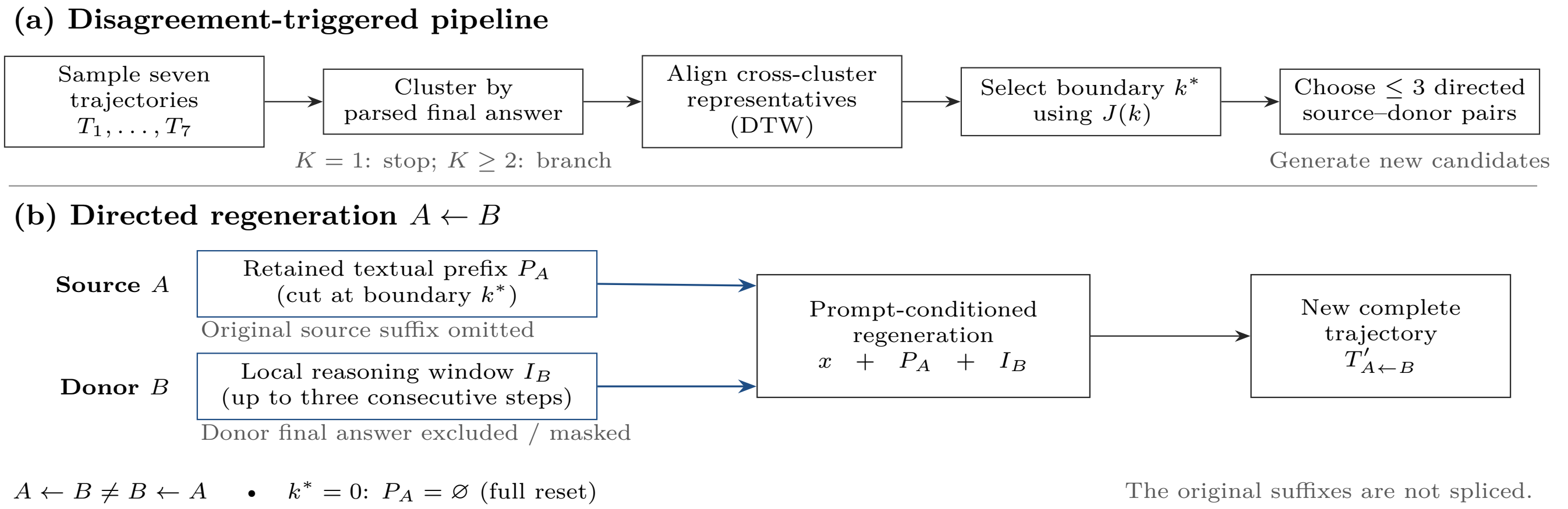
\begin{figure*}[t]
  \centering
  \resizebox{\textwidth}{!}{\input{figures/asb_overview.tex}}
  \caption{Adaptive Semantic Branching. (a) The inference pipeline groups seven independent trajectories by answer, aligns cross-cluster representatives, locates the persistent divergence boundary, and selects directed exchanges. (b) In $A\leftarrow B$, source prefix $P_A$ and donor insight $I_B$ condition a newly generated trajectory $T'_{A\leftarrow B}$. The $k^*=0$ case uses an empty source prefix and a fresh start.}
  \label{fig:operator}
\end{figure*}

We propose \emph{Adaptive Semantic Branching} (\asb), an inference-time framework that makes these decisions from completed reasoning traces. The method begins with seven independently sampled trajectories. Their extracted answers determine clusters, and representatives of different clusters are compared through step-level monotonic alignment. A change-point objective rewards an earlier region of higher cross-trajectory similarity followed by a sustained decrease in similarity. The resulting boundary defines a textual prefix from a source trajectory. A donor trajectory provides a short, locally aligned reasoning window. Conditioning on the original problem, source prefix, and donor insight yields a new candidate trajectory (Figure~\ref{fig:operator}).

The direction of exchange matters. $A\leftarrow B$ retains the reasoning prefix from $A$ and uses a local alternative from $B$, whereas $B\leftarrow A$ constructs a different prompt from a different prefix and insight window. \asb\ evaluates these directed choices, uses a divergence-based score to rank candidate exchanges, and allocates zero, two, or three branch generations according to the number of answer clusters. The resulting procedure is lightweight: step features are computed locally, while the initial and branch generations use the same underlying language model.

Our study develops a concrete \emph{cross-trajectory regeneration operator} and reports its first candidate-coverage evaluation. On the fixed 500-item MATH manifest, the seven initial samples contain a correct answer for 445 problems. Directed regeneration expands that coverage to 455 problems, recovering 18.18\% of the cases missed by the initial pool. In addition to the empirical result, this paper details the complete sequence from raw reasoning segmentation to branch lineage, deterministic seed derivation, and generation prompts.

The contributions are threefold:
\begin{itemize}
  \item \textbf{Divergence-localized branching.} A monotonic step alignment and persistent similarity-change objective produce a source-specific regeneration boundary from conflicting completed trajectories.
  \item \textbf{Directed cross-trajectory regeneration.} A source prefix and donor-local insight form a compact conditioning context for a fresh reasoning trajectory, with explicit directionality and traceable parent lineage.
  \item \textbf{Adaptive candidate expansion.} Answer-cluster disagreement determines a small branch budget, yielding a reproducible generation policy and a measured improvement in oracle candidate coverage on MATH-500.
\end{itemize}

\section{Related Work}
\paragraph{Test-time sampling and structured exploration.}
Self-consistency \cite{wang2023selfconsistency} uses multiple independently sampled chains to aggregate predictions. Tree of Thoughts \cite{yao2023tot} and Graph of Thoughts \cite{besta2024got} develop structured reasoning spaces with intermediate exploration and composition. \asb\ uses the structure already present across \emph{completed} samples: the final-answer clusters determine which trajectories to compare, and aligned intermediate steps determine how new candidate generations are conditioned.

\paragraph{Information exchange across reasoning paths.}
Multi-Chain Reasoning \cite{yoran2023mcr} integrates information from multiple sampled chains through meta-reasoning. Contract-checked graph editing \cite{li2026graphs} constructs combinations of partial solutions represented as structured reasoning graphs. \asb\ introduces a complementary natural-language operation: a directionally chosen textual prefix and a short, boundary-localized donor window are assembled into a new generation prompt. The operation can be applied directly to free-form chain-of-thought outputs.

\paragraph{Divergence and selective intervention.}
Reasoning Path Divergence \cite{ju2025rpd} develops step-level trajectory comparison for measuring diversity. AgentAuditor \cite{yang2026auditor} analyzes agreements and disagreements within multi-agent reasoning trees for localized adjudication. Entropy-Gated Branching \cite{li2026egb} uses uncertainty to allocate branch computation during generation, while Trap-Aware Adaptive Restart \cite{chen2026taar} uses a learned diagnostic policy to restart reasoning from identified prefixes. \asb\ connects completed-path step alignment to a persistent-divergence boundary and a directed candidate-generation policy. Its branch location emerges from the \emph{relationship between two solutions}, rather than from a single trajectory's uncertainty signal.

\section{Problem Formulation}
For problem $x$, sample $N$ independent model trajectories $T_1,\ldots,T_N$ from $p_\theta$. Each trajectory contains the original reasoning text $r_i$ and a parsed final answer $a_i$. In the reference implementation $N=7$. Let $\kappa(a_i)$ be the extracted-answer clustering key, producing ordered answer clusters $\mathcal{G}_1,\ldots,\mathcal{G}_K$.

The method constructs a collection of new branch trajectories $\mathcal{B}(x)$ and returns
\begin{equation}
  \mathcal{C}(x)=\mathcal{C}_0(x)\cup\mathcal{B}(x),\qquad
  \mathcal{C}_0(x)=\{T_1,\ldots,T_7\},
\end{equation}
with up to three additional trajectories per problem. The primary evaluation quantity is \emph{oracle candidate coverage}: the fraction of problems for which the returned collection contains a gold-equivalent answer. Let $y_q$ be the reference answer to problem $x_q$ and let $\operatorname{Eq}$ be the MATH equivalence checker. Then
\begin{equation}
\label{eq:coverage}
\cove(\mathcal{C})=\frac{1}{M}\sum_{q=1}^{M}
\mathbf{1}\!\left[\exists\, T\in\mathcal{C}(x_q):\operatorname{Eq}(a_T,y_q)=1\right].
\end{equation}
This metric quantifies how many problems have at least one correct candidate available for a subsequent selection stage. The reference implementation evaluates $\operatorname{Eq}$ after generation and uses the parsed answer string as the inference-time cluster key.

\section{Adaptive Semantic Branching}
\subsection{Initial generation and answer clusters}
For every problem, \asb\ generates seven complete trajectories with temperature $0.7$, top-$p=1.0$, and maximum completion length 2,048 tokens. Each call receives an independently derived deterministic seed. The initial prompt requests a complete derivation and a boxed final answer. The seven calls use the same problem statement and are sampled independently.

The extracted answers partition the trajectories into clusters ordered by their first appearance. When $K=1$, \asb\ finishes with the initial seven trajectories. When $K\ge2$, the method selects one representative trajectory from each cluster. With token log-probabilities available, the representative has the largest length-normalized mean token log-probability; otherwise the first observed trajectory is selected. All ordered cross-cluster representative pairs form candidates for directed regeneration.

\subsection{Reasoning-step extraction}
Let $r_A$ be the raw reasoning text of a representative trajectory. \asb\ divides it into ordered steps $A=(a_1,\ldots,a_m)$, storing the start and end character offsets of every step. The segmentation procedure gives priority to numbered lists and bullet formatting, using their line boundaries as steps. For other response formats, line breaks and sentence punctuation supply candidate boundaries. A final answer-only fragment, such as a standalone boxed answer, is excluded from the reasoning sequence. Segments shorter than five whitespace-delimited tokens are merged into the preceding step.

The retained offsets serve a concrete role in regeneration: once a step boundary is selected, the source prefix is recovered directly from the original response text. The implementation also records the step index, text, and token count for inspection and analysis.

\subsection{Deterministic step representation}
The MATH-500 reference run employs a local \emph{word- and character-gram hashing encoder} of dimension $D=512$. For a step $a$, lowercase its text, split it on whitespace, and rejoin the words with single spaces. The feature multiset $F(a)$ contains all whitespace-delimited words and all character substrings of length four from the rejoined string.

For each feature $f$, compute its SHA-256 digest $h(f)$. The first four digest bytes, interpreted as an unsigned integer, select the coordinate $q(f)=\operatorname{int}(h(f)_{0:4})\bmod 512$. The least significant bit of the fifth byte determines the sign $\sigma(f)\in\{-1,+1\}$. The unnormalized vector is
\begin{equation}
  v_a[d]=\sum_{f\in F(a)}\sigma(f)\mathbf{1}[q(f)=d],
  \quad d=0,\ldots,511.
\end{equation}
The embedding is $e(a)=v_a/\lVert v_a\rVert_2$ for a nonzero vector and the zero vector otherwise. Cosine similarity involving a zero vector is defined as zero. This fixed-dimensional representation produces deterministic step similarities using local string operations and hashing.

\subsection{Monotonic cross-trajectory alignment}
For a directed representative pair $A\leftarrow B$, write source steps $A=(a_1,\ldots,a_m)$ and donor steps $B=(b_1,\ldots,b_n)$. The pairwise cost is
\begin{equation}
  c(i,j)=1-\cos\!\left(e(a_i),e(b_j)\right).
\end{equation}
A dynamic time warping (DTW) program \cite{sakoe1978dtw} finds a minimum-cost, order-preserving alignment path $\pi$. For source step $a_i$, aggregate the similarity of its aligned donor steps by
\begin{equation}
  s_i=\max_{j:(i,j)\in\pi}\cos\!\left(e(a_i),e(b_j)\right).
\end{equation}
The implementation fills any degenerate unmatched source position with the donor step at the corresponding proportional position. The result $s_1,\ldots,s_m$ describes how agreement between the two reasoning sequences changes over the progress of the source path.

\subsection{Persistent divergence localization}
For each interior source boundary $k\in\{1,\ldots,m-1\}$, \asb\ computes
\begin{equation}
\label{eq:j}
J(k)=\frac{\sqrt{k(m-k)}}{m}\left(\frac{1}{k}\sum_{i=1}^{k}s_i-\frac{1}{m-k}\sum_{i=k+1}^{m}s_i\right).
\end{equation}
The first term rewards a reasonably balanced split, while the difference in average similarities favors a relatively aligned prefix followed by a persistently divergent suffix. \asb\ selects the earliest maximizing boundary when its score is positive. For a nonpositive maximum, the selected boundary is $k^*=0$, initiating a full-reset collaborative branch. This allows the donor direction to be explored even for pairs whose similarity profile favors restarting from the original problem.

For $k^*>0$, let $d_A(k^*)$ denote the original character offset at the end of source step $k^*$. The source prefix is recovered as
\begin{equation}
  P_A=r_A[\,0:d_A(k^*)\,].
\end{equation}
At $k^*=0$, $P_A$ is the empty string. The metadata records the step-based prefix ratio $k^*/m$ together with the character offset, allowing both the chosen reasoning boundary and the corresponding raw text to be reconstructed.

\subsection{Donor-local insight and directed regeneration}
The donor step aligned to the chosen source boundary determines the start of an insight window $I_B$. \asb\ extracts that donor step and up to two subsequent donor steps, for a maximum window length of three steps. The donor's final-answer field is kept in lineage metadata; the branch prompt contains the partial insight text with any incidental final-answer strings masked.

A directed branch conditions the same model on the original problem, source prefix, and donor insight:
\begin{equation}
\label{eq:operator}
T'_{A\leftarrow B}\sim p_\theta\!\left(\cdot\mid
\operatorname{Prompt}(x,P_A,I_B)\right).
\end{equation}
The generation instruction asks the model to check the competing directions, develop its own continuation, and return a complete derivation with a boxed answer. The prefix is supplied as text inside the user prompt. A branch with $k^*=0$ presents the alternative donor insight alongside the original problem and requests a solution from the beginning. Each branch becomes an additional candidate with a full parent--donor lineage record.

The direction $A\leftarrow B$ determines both the retained prefix and the borrowed insight, so reversing the pair creates a distinct search operation. Figure~\ref{fig:operator}(b) depicts this asymmetric transfer.

\subsection{Pair ranking and adaptive branch budget}
For a directed pair, the divergence strength is $J_{\max}=\max_k J(k)$. When token log-probabilities are stored, the mean log-probabilities for the source-prefix span and the donor-window span are exponentiated to obtain local confidence factors $c_A$ and $c_B$. The branch score is
\begin{equation}
  q(A\leftarrow B)=J_{\max}\sqrt{c_Ac_B}.
\end{equation}
In the absence of token log-probabilities, the confidence factors take value one and pair ranking uses $J_{\max}$ directly.

The branching policy follows the answer-cluster structure. For $K=1$, no additional generations are allocated. For $K=2$, both directed pairs are generated, yielding two branches. For $K\ge3$, candidates are ordered by branch score and up to three are chosen, prioritizing different unordered cluster pairs and diversity among sources and donors. Additional ranked pairs fill any remaining slots. The complete candidate collection therefore contains 7, 9, or up to 10 generated trajectories per problem.

\begin{algorithm}[t]
\caption{Adaptive Semantic Branching (per problem)}
\label{alg:asb}
\small
\begin{algorithmic}[1]
\Require Problem $x$, model $p_\theta$, budget $B=3$
\State $\mathcal{C}_0\gets\operatorname{SampleIndependent}(x,7)$
\State $\mathcal{G}\gets\operatorname{ClusterByAnswer}(\mathcal{C}_0)$
\If{$|\mathcal{G}|=1$}\State \Return $\mathcal{C}_0$ \EndIf
\State $R\gets\operatorname{Representatives}(\mathcal{G})$
\For{each ordered cross-cluster pair $(A,B)$ from $R$}
  \State $s,\pi\gets\operatorname{AlignSteps}(A,B)$
  \State $k^*\gets\operatorname{PersistentCut}(s)$ using Eq.~\ref{eq:j}
  \State $P_A\gets\operatorname{RawPrefix}(A,k^*)$
  \State $I_B\gets\operatorname{DonorWindow}(B,\pi,k^*)$
  \State $q(A\leftarrow B)\gets\operatorname{BranchScore}(A,B,k^*)$
\EndFor
\State $\mathcal{P}\gets\operatorname{ChoosePairs}(R,q,B)$
\State $\mathcal{B}\gets\varnothing$
\For{each $(A,B)$ in $\mathcal{P}$}
  \State $T'\gets\operatorname{Generate}(x,P_A,I_B)$
  \State $\mathcal{B}\gets\mathcal{B}\cup\{T'\}$
\EndFor
\State \Return $\mathcal{C}_0\cup\mathcal{B}$
\end{algorithmic}
\end{algorithm}

\section{Experimental Setup}
\paragraph{Dataset and model.}
We evaluate the reference configuration on a fixed, ordered manifest of 500 MATH problems, following the established MATH benchmark \cite{hendrycks2021math,lightman2023verify}. The model is Qwen2.5-7B-Instruct \cite{yang2024qwen}. All initial and branch calls use temperature $0.7$, top-$p=1.0$, and maximum completion length 2,048 tokens. The base seed is 100, with independent identifier-derived seeds for the seven initial trajectories and each new branch.

\paragraph{Evaluation and accounting.}
Candidate coverage is computed with the MATH answer-equivalence checker (Eq.~\ref{eq:coverage}). Inference-time answer clusters are formed from parsed answer strings. We report absolute problem counts, coverage percentages, newly recovered problems, and model generation calls. Each extra branch consumes one model generation call; the local embedding and alignment computations operate on stored trajectory text. The analysis uses the recorded seed-100 run described by the implementation specification.

\section{Results}
\begin{table}[t]
\centering
\caption{Oracle candidate coverage and generation counts for the stored MATH-500 run (Qwen2.5-7B-Instruct; base seed 100).}
\label{tab:results}
\small
\begin{tabular}{@{}p{0.63\columnwidth}r@{}}
\toprule
Metric & Value \\
\midrule
Initial candidate coverage & $445/500$ (89.0\%) \\
\asb\ candidate coverage & $455/500$ (91.0\%) \\
Additional problems recovered & $10/500$ (+2.0 pp) \\
Recovery among 55 initial misses & $10/55$ (18.18\%) \\
Initial model generation calls & 3,500 \\
Additional branch calls & 576 \\
Total model generation calls & 4,076 \\
Average calls per problem & 8.152 \\
Problems with 0 / 2 / 3 branches & 284 / 72 / 144 \\
\bottomrule
\end{tabular}
\end{table}

\paragraph{Candidate expansion.}
The initial seven-sample collection contains a correct trajectory for 445 of the 500 problems, yielding 89.0\% oracle coverage. After the directed branches are added, the expanded collection covers 455 problems, or 91.0\%. The 10 newly recovered problems correspond to an absolute increase of 2.0 percentage points and recovery of $10/55=18.18\%$ of the problems initially missing a correct candidate. These gains measure the enlarged solution space available to answer-selection methods.

\paragraph{Adaptive compute allocation.}
The branch policy responds to the observed answer-cluster count. For 284 problems (56.8\%), the seven initial trajectories share one parsed-answer cluster, so the method retains the initial collection. For 72 problems (14.4\%), two answer clusters trigger two directed exchanges. The remaining 144 problems (28.8\%) receive three branches. The resulting total is $72\times2+144\times3=576$ additional calls, corresponding to 1.152 extra calls per problem and an average of 8.152 calls per problem overall. Relative to the seven-call starting point, the generation-call budget increases by 16.46\%.

\paragraph{Observed behavior.}
The key outcome is the appearance of gold-equivalent answers in trajectories generated \emph{after} inspecting disagreement between the original candidates. The branch lineage and boundary metadata make these recoveries analyzable at the level of individual source--donor exchanges, prefix locations, and donor windows. This creates a direct experimental interface for studying how local differences between reasoning paths translate into new candidate solutions.

\section{Discussion and Evaluation Directions}
The central design choice in \asb\ is to place collaboration at the transition between related and diverging reasoning paths. The method uses a balanced change-point criterion to localize this transition, a short aligned donor window to introduce a competing direction, and an asymmetric prefix-conditioned operator to construct the next candidate. The result is a compact, auditable inference workflow with an explicit connection between disagreement structure and generation budget.

The reference implementation also makes several natural extensions straightforward. A matched-call independent sampling comparison will measure the value of cross-trajectory conditioning. Random boundary selection, prefix-only generation, and donor-only generation will isolate the roles of localization and information transfer. Alternative step encoders and answer-equivalence-based clustering can test sensitivity to the lexical representation and answer formatting. Repeated seeds, additional models, and a shared candidate selector will extend the evaluation from candidate discovery to downstream answer selection. Each extension reuses the stored lineage and step-alignment records.

\section{Conclusion}
We presented Adaptive Semantic Branching, a training-free method that converts disagreements between completed reasoning trajectories into directed, localized regeneration. Its core operation aligns source and donor reasoning steps, identifies a persistent divergence boundary, preserves a source-specific textual prefix, and introduces a short competing reasoning window to generate a new solution. Answer-cluster structure determines how a small branch budget is allocated. On a stored MATH-500 run with Qwen2.5-7B-Instruct, the method expands correct-candidate coverage from 89.0\% to 91.0\%, recovering 10 initially missed problems with 576 additional generations across 500 questions. The full procedure supplies a reproducible basis for studying divergence-guided candidate generation at test time.

\appendix
\section{Reproducibility Specification}
\label{sec:repro}
\paragraph{Reference configuration.}
Table~\ref{tab:config} collects the parameters used in the reported run. The evaluation manifest preserves its original 500-item order; each problem and each generated branch receives its own derived seed. Parsing, embedding, alignment, and scoring operate on the completed responses before branch prompts are formed.

\begin{table}[h]
\centering
\caption{Reference implementation settings for the MATH-500 seed-100 run.}
\label{tab:config}
\small
\begin{tabular}{@{}ll@{}}
\toprule
Parameter & Configuration \\
\midrule
Model & Qwen2.5-7B-Instruct \\
Benchmark & fixed MATH-500 manifest \\
Initial trajectories & 7 per problem \\
Maximum branches & 3 per problem \\
Temperature / top-$p$ & 0.7 / 1.0 \\
Max completion tokens & 2,048 \\
Base seed & 100 \\
Branch seed mode & independent per branch \\
Embedding dimension & 512 \\
Embedding features & words + character 4-grams \\
Alignment & monotonic DTW \\
Clustering key & parser-extracted answer \\
Coverage evaluator & MATH answer equivalence \\
\bottomrule
\end{tabular}
\end{table}

\paragraph{Stable seed derivation.}
The seed function builds the UTF-8 string
\begin{lstlisting}
import hashlib
key = f"{BASE_SEED}|{sample_id}|{stage}|{agent_or_branch_id}"
seed = int.from_bytes(
    hashlib.sha256(key.encode("utf-8")).digest()[:8], "big"
) % (2**31 - 1)
\end{lstlisting}
Here \texttt{BASE\_SEED=100}. Initial calls have distinct agent identifiers and stage names, while branches use \texttt{stage="branch"} and their branch identifier. The implementation also supports a source-seed experimental setting; the reported configuration uses independently derived branch seeds.

\paragraph{Step offsets and embedding details.}
The step extractor records source-text character spans before computing embeddings. Its precedence is numbered/bulleted lines, then line breaks and punctuation, with final-answer-only fragments removed and fragments shorter than five whitespace tokens joined to the preceding step. The SHA-256 word/character feature encoder uses the first four digest bytes for the signed-hashing coordinate and the fifth byte for its sign. The final vectors are normalized in $\ell_2$ and their cosine values enter the DTW cost matrix. The selected prefix is reconstructed using the original character offset $d_A(k^*)$.

\paragraph{Initial generation prompt.}
The reference run formats the following content as Qwen chat messages. The complete template is reproduced here with placeholders for the problem statement:
\begin{lstlisting}
<|im_start|>system
Solve the mathematics problem independently and carefully.
Show a complete derivation and put the final answer in \boxed{}.
Do not use any external answer or majority information.<|im_end|>
<|im_start|>user
{problem}<|im_end|>
<|im_start|>assistant
\end{lstlisting}

\paragraph{Branch generation prompt.}
For every selected directed pair, the implementation constructs the following prompt with the original problem, source prefix, and donor-local insight:
\begin{lstlisting}
<|im_start|>system
Solve the mathematics problem carefully. Show a complete derivation and put the final answer in \boxed{}.<|im_end|>
<|im_start|>user
Problem:
{problem}

{preserved_source_prefix_or_no_reliable_prefix_message}

Alternative partial reasoning direction (no final answer):
{donor_insight_window}

Continue solving the original problem. Use the alternative only as a competing hypothesis; do not copy it or assume it is correct. Explicitly check where the approaches diverge, derive the continuation yourself, and return a complete solution with the final answer in \boxed{}.
<|im_end|>
<|im_start|>assistant
\end{lstlisting}
For $k^*>0$, the source block begins with \texttt{Preserved source reasoning prefix (exact):}, followed by the exact extracted source prefix and an instruction to retain it unless logically inconsistent. For $k^*=0$, the source block requests a fresh solution that considers the competing partial direction. The donor window contains up to three aligned consecutive steps, and any occurrences of the donor final-answer string are masked during prompt construction.

\paragraph{Run artifacts and lineage.}
The experiment produces \texttt{run\_config.json}, \texttt{results.jsonl}, \texttt{summary.json}, and \texttt{gold\_recovery\_cases.jsonl}. Each stored initial trajectory includes prompt and response text, parsed answer, sampling seed, finish reason, and token metadata when supplied by the generation backend. Each branch additionally stores its source and donor trajectory identifiers, answer clusters, divergence step, raw offset, $J_{\max}$, prefix ratio, extracted donor insight, branch score, seed, prompt, and generated response. These fields support reconstruction of the pair-selection decisions and inspection of newly recovered answers.

\bibliographystyle{plainnat}
\bibliography{references}
\end{document}

%% file: figures/asb_overview.tex
\begin{tikzpicture}[
    x=1cm,y=1cm,>=Stealth,
    every node/.style={font=\scriptsize,inner sep=1pt},
    stage/.style={draw=black!80,rectangle,align=center,minimum height=0.70cm,text width=2.62cm,inner sep=3pt},
    selected/.style={draw=deepblue,rectangle,align=center,minimum height=0.66cm,text width=4.15cm,inner sep=3pt},
    promptbox/.style={draw=black!85,rectangle,align=center,minimum height=1.34cm,text width=3.35cm,inner sep=4pt},
    resultbox/.style={draw=black!85,rectangle,align=center,minimum height=1.34cm,text width=2.55cm,inner sep=4pt},
    link/.style={-Stealth,draw=black!85,thin},
    branchlink/.style={-Stealth,draw=deepblue,semithick},
    desc/.style={font=\scriptsize,align=left,text=black!65}
]
\node[anchor=west,font=\small\bfseries] at (0,4.30) {(a) Disagreement-triggered pipeline};
\node[stage] (sample) at (1.37,3.42) {Sample seven\\trajectories $T_1,\ldots,T_7$};
\node[stage] (clusters) at (4.85,3.42) {Cluster by\\parsed final answer};
\node[stage] (alignment) at (8.33,3.42) {Align cross-cluster\\representatives (DTW)};
\node[stage] (cut) at (11.81,3.42) {Select boundary $k^*$\\using $J(k)$};
\node[stage] (budget) at (15.29,3.42) {Choose $\leq 3$ directed\\source--donor pairs};
\foreach \from/\to in {sample/clusters,clusters/alignment,alignment/cut,cut/budget}{
  \draw[link] (\from.east) -- (\to.west);
}
\node[desc,anchor=north,align=center] at (4.85,2.91) {$K=1$: stop; $K\geq2$: branch};
\node[desc,anchor=north,align=center] at (15.29,2.91) {Generate new candidates};
\draw[black!45,thin] (0,2.50) -- (16.68,2.50);

\node[anchor=west,font=\small\bfseries] at (0,2.16) {(b) Directed regeneration $A\leftarrow B$};

\node[anchor=east,font=\scriptsize\bfseries] at (1.80,1.43) {Source $A$};
\node[selected,anchor=west] (prefix) at (2.05,1.43)
   {Retained textual prefix $P_A$\\(cut at boundary $k^*$)};
\node[desc,anchor=west] at (2.05,0.91) {Original source suffix omitted};

\node[anchor=east,font=\scriptsize\bfseries] at (1.80,0.31) {Donor $B$};
\node[selected,anchor=west] (insight) at (2.05,0.31)
   {Local reasoning window $I_B$\\(up to three consecutive steps)};
\node[desc,anchor=west] at (2.05,-0.21) {Donor final answer excluded / masked};

\node[promptbox,anchor=west] (prompt) at (8.16,0.86)
   {Prompt-conditioned\\regeneration\\$x\; +\; P_A\; +\; I_B$};
\node[resultbox,anchor=west] (out) at (13.55,0.86)
   {New complete trajectory\\$T'_{A\leftarrow B}$};

\draw[branchlink] (prefix.east) -- ([yshift=0.55cm]prompt.west);
\draw[branchlink] (insight.east) -- ([yshift=-0.55cm]prompt.west);
\draw[link] (prompt.east) -- (out.west);

\node[anchor=west,font=\scriptsize] at (0,-0.85)
  {$A\leftarrow B\neq B\leftarrow A$ \quad\textbullet\quad $k^*=0$: $P_A=\varnothing$ (full reset)};
\node[desc,anchor=east] at (16.68,-0.85)
  {The original suffixes are not spliced.};
\end{tikzpicture}